# ELECTROOPTICAL IMAGE SYNTHESIS FROM SAR IMAGERY USING GENERATIVE ADVERSARIAL NETWORKS

**Grant Rosario**

PeopleTec, Inc.

Huntsville, AL

grant.rosario@peopletec.com

**David A. Noever**

PeopleTec, Inc.

Huntsville, AL

david.noever@peopletec.com

## ABSTRACT

*The utility of Synthetic Aperture Radar (SAR) imagery in remote sensing and satellite image analysis is well established, offering robustness under various weather and lighting conditions. However, SAR images, characterized by their unique structural and texture characteristics, often pose interpretability challenges for analysts accustomed to electrooptical (EO) imagery. This application compares state-of-the-art Generative Adversarial Networks (GANs) including Pix2Pix, CycleGan, S-CycleGan, and a novel dual-generator GAN utilizing partial convolutions and a novel dual-generator architecture utilizing transformers. These models are designed to progressively refine the realism in the translated optical images, thereby enhancing the visual interpretability of SAR data. We demonstrate the efficacy of our approach through qualitative and quantitative evaluations, comparing the synthesized EO images with actual EO images in terms of visual fidelity and feature preservation. The results show significant improvements in interpretability, making SAR data more accessible for analysts familiar with EO imagery. Furthermore, we explore the potential of this technology in various applications, including environmental monitoring, urban planning, and military reconnaissance, where rapid, accurate interpretation of SAR data is crucial. Our research contributes to the field of remote sensing by bridging the gap between SAR and EO imagery, offering a novel tool for enhanced data interpretation and broader application of SAR technology in various domains.*

## KEYWORDS

*Geospatial Data, Image Generator, Machine Learning, Synthetic,*

# 1. INTRODUCTION

Synthetic Aperture Radar (SAR) systems are capable of creating high-resolution remote sensing images of the earths surface from satellite and aircraft. These images offer several key advantages over standard electro-optical (EO) images, most significantly, the ability to penetrate clouds and operate independently of daylight, which has led to SAR systems being deployed extensively in various fields, including environmental monitoring, natural disaster assessment, military reconnaissance, and geological mapping [1]. Figure 1 shows the benefit of a SAR image when cloud coverage is present.

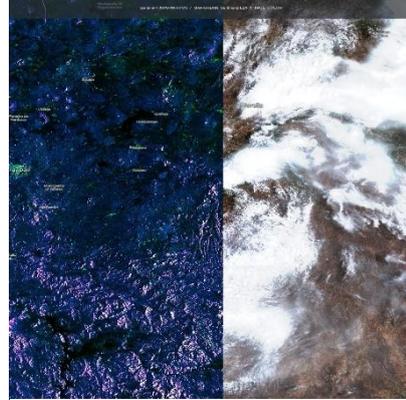

*Figure 1: SAR image on left compared with cloud covered optical image on right.*

Despite these advantages, SAR images poses significant challenges and still has drawbacks compared to EO images, specifically regarding human interpretability. The radar signals used to create SAR images result in image data that is fundamentally different from the visual information captured by traditional optical sensors. SAR images can often appear dark, noisy, and contain speckle patterns that can obscure important features. This inherent complexity makes it difficult for non-expert users to analyse and understand SAR data effectively.

EO imagery, on the other hand, provides visual representations that are more intuitive and easier for humans to interpret. These images capture the reflected light from the Earth's surface in various spectral bands, producing images that closely resemble what the human eye perceives. This makes EO imagery highly valuable for applications that require detailed visual analysis, such as urban planning, agricultural monitoring, and infrastructure assessment. Figure 2 compares a SAR image with a clear EO image, showing the difference and advantage in clarity that EO provides.

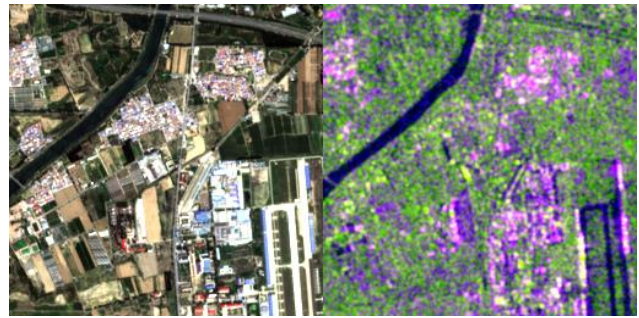

*Figure 2: EO image on left compared with SAR counterpart on right.*

This creates a potential area of maximum benefit where both SAR and EO images can both be utilized for each other's strengths. This is the primary motivation behind the growing field of research focused on SAR-to-EO translation. Generative Adversarial Networks (GANs), introduced by Goodfellow et al. in 2014 [2], have revolutionized the field of image synthesis by enabling the generation of high-quality, realistic images from various input types. By employing GANs for SAR-to-EO image translation, we can create EO-like images from SAR data, enhancing the interpretability and usability of SAR imagery.

In this paper, we present two novel solutions to problems within the field of SAR-to-EO image translation. The first is a novel GAN architecture designed to improve upon the existing state-of-the-art (SOTA) models capable of translating SAR images to EO by utilizing noise de-speckling as well as transformer methodology. The second is a technique for providing visualizations and metrics to the end user in order to improve confidence in the model translations. While there are many new models being developed to improve the accuracy of SAR-to-EO translations, we have yet to see a solution for providing the end user with a metric so they can engage confidently with the translated image.

## 2. CURRENT SAR-TO-EO APPROACHES

**Current Translation Approaches**

Numerous approaches have been developed to address the challenge of translating SAR images to EO images. Traditional methods often relied on physical modeling and statistical techniques, which, while effective in certain contexts, often failed to capture the complex, non-linear relationships between SAR and EO imagery. With the advent of deep learning, particularly convolutional neural networks (CNNs), more sophisticated and capable models have emerged.

Early attempts using CNNs focused on direct translation models, which aimed to learn a mapping from SAR to EO images by minimizing pixel-wise differences. These methods, however, struggled with issues such as texture detail loss and the inability to generate realistic, high-resolution EO images. To overcome these challenges, researchers began to explore adversarial training frameworks, which have proven to be highly effective in various image synthesis tasks.

This section surveys some of the most recent existing methods for SAR to EO image translation. New models are constantly being researched and developed; therefore, we chose to focus on four unique models which utilize key technologies such as Generative Adversarial Networks (GANs). The models we compared are Pix2Pix [3], S-CycleGAN [4], SAR2EO [5], and Dual-Generator [6]. The following sections will detail the structure and functionality of each model as well as highlight the shortcomings and gaps in their approaches. Lastly, we will end this section by introducing our own model which combines some key functionality of the compared models to arrive at an optimal method for translation.

**Pix2Pix**

One of the pioneering works in this area is the Pix2Pix model, a conditional GAN that learns a mapping from input images to output images and simultaneously learns a loss function to train this mapping. Introduced by Isola et al. (2017), Pix2Pix consists of a generator that produces new images in the desired format and a discriminator that differentiates between real images in the target format and those generated by the generator. A few examples provided in the paper are transforming semantic labels to an optical street scene or a grayscale image to color. [3] The method for implementing Pix2Pix consists of first augmenting the input data through random cropping, resizing, flipping, and normalizing followed by feeding the augmented data into the generator. The generator consists of a U-Net architecture structure where the encoder and decoder within the generator have a shared data stream via skip connections among each layer. Figure 3 shows a visualization of the U-Net architecture from the authors of the original paper [7]. The generated output is then passed into a discriminator with a proposed PatchGAN architecture in order to focus the model attention on local image patch structure. The discriminator tries to classify if each patch of the generated image is real or fake by computing an adversarial loss which pushes the generator to produce images that are indistinguishable from ground truth images, while the L1 loss encourages the generated images to be close to the ground truth in a pixel-wise sense [3].

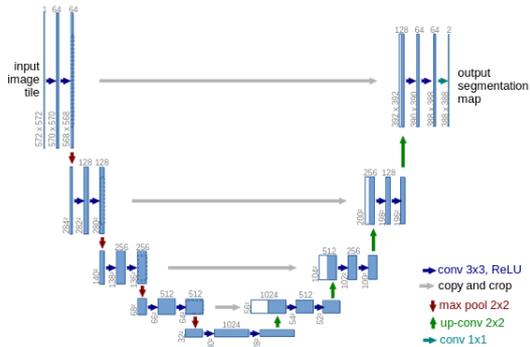

*Figure 3: U-net architecture introduced in Ronneburger et al's 2015 work on ConvNets.*

Despite its success, Pix2Pix often suffers from generating blurry images when dealing with high-resolution data (shown in Table 1) which led to a 2021 paper by Zuo et al. proposing an updated

Pix2Pix model for SAR-to-Optical image translation which uses a phase consistency constraint [8]. For our comparisons, we kept the Pix2Pix implementation as close to the original paper as possible.

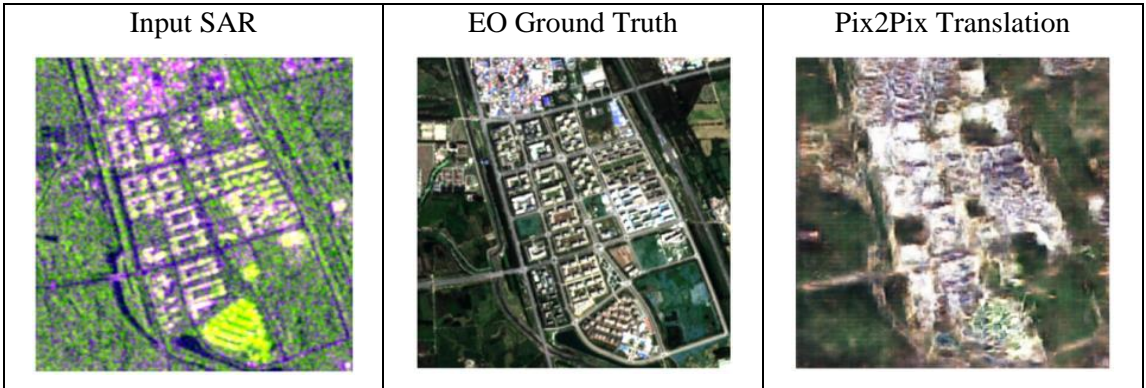

Table 1: Translated SAR image using Pix2Pix model.

**S-CycleGAN**

To address the shortcomings of direct translation models like Pix2Pix, CycleGAN, proposed by Zhu et al. (2017), and its variants, including S-CycleGAN, have been developed with the benefit of not requiring direct translated counterparts of original input images for training [9]. For example, if we're training a model to translate paintings into photorealistic images, we do not need photorealistic versions of the exact same photos as the training data for the model to learn to translate the styles. CycleGAN accomplishes this by introducing the concept of cycle consistency to ensure that the translated image can be mapped back to the original image. S-CycleGAN, a specific variant tailored for SAR to EO translation, enhances this framework by incorporating structural consistency loss to better preserve the spatial structures inherent in SAR images while generating EO images. This model comprises two generators and two discriminators: one generator translates SAR images to EO images, and the other translates EO images back to SAR. The cycle consistency loss ensures that translating an image to the other domain and back results in the original image, while the adversarial loss ensures the realism of the generated images [4]. Table 2 shows the result of a translated optical image using a trained S-CycleGan model.

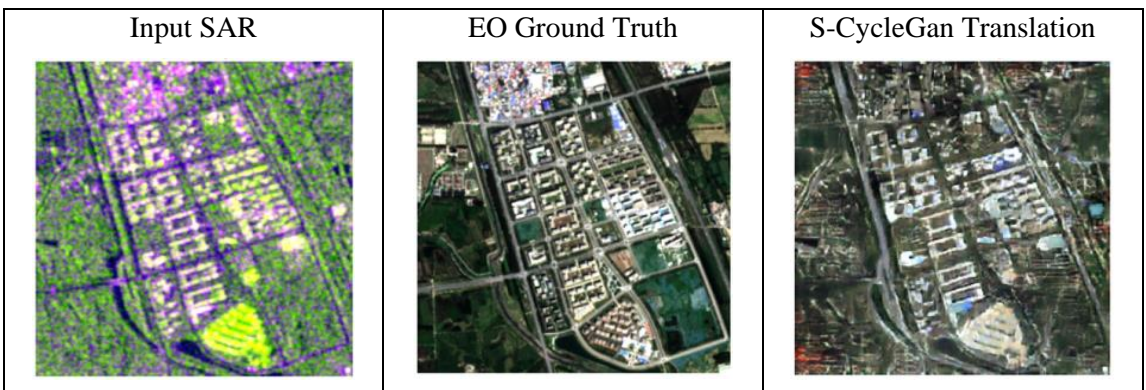

Table 2: Translated SAR image using S-CycleGan model.

**SAR2EO**

As stated previously, one of the drawbacks of SAR images versus their EO counterparts is the amount of noise and speckling that appears in the image. In order to address this, there have been promising proposed methods of de-noising or de-speckling SAR images [10] [11]. This provides the potential to use a less noisy SAR image as input into a translation model. Shenshen Du et al.

proposed their 2023 SAR2EO model which employed a denoising algorithm as an augmentation for the SAR images before they are fed into the generator. The translation pipeline then utilized a Pix2PixHD inspired model which employed two generators and two discriminators, one for a high-resolution translation and one for a low-resolution translation with the intention being the high res would focus on more local features while the low-res would focus on global features [12]. Table 3 shows the SAR2EO translation result of our baseline SAR image based on our attempted reconstruction of the SAR2EO architecture.

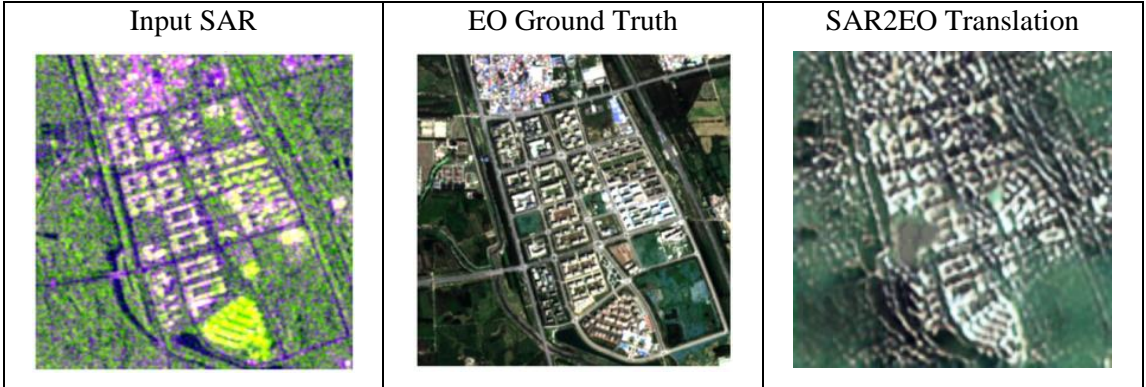

*Table 3: Translated SAR image using SAR2EO model.*

**Dual-Generator SAR-to-EO Translator**

Nie et al. offered a novel solution to some of the drawbacks of these previous models in their 2022 paper proposing a SAR-to-EO translation network which, similarly to the previously discussed models, utilized dual generators, but uniquely implemented them to focus solely on SAR structure and texture features rather than dedicate one of them to the EO image. They further expounded their model by adding a Bidirectional Gated Feature Fusion (Bi-GFF) module and a Contextual Feature Aggregation (CFA) module in order to fuse the features and refined the generated output image. The generated image is then fed into a discriminator along with its edge features to detect whether it is real or fake, thus providing the generator with the motivation to improve the generated image [6]. An example output of the dual-generator model is shown in Table 4.

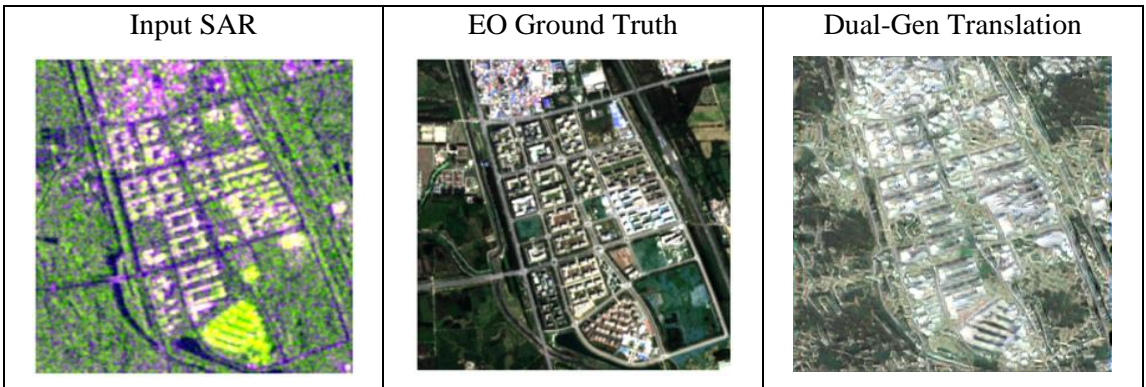

*Table 4: Translated SAR image using dual-generator model.*

We believe this models use of texture and structure isolation combined with feature fusion and the transformer-like functionality of the CFA modules provided the most promising generated output, so we used this model as our baseline while proposing our own iteration that we believe offers improvements to the model output.

## 2. SHORTCOMINGS AND GAPS IN CURRENT APPROACHES

While the advancements in GAN-based models have significantly improved SAR to EO image translation, several challenges and gaps remain.

**2.1 Resolution and Detail Preservation**

One of the primary challenges is the generation of high-resolution and detailed EO images. Models like Pix2Pix and even S-CycleGAN, although better, often produce images that lack fine details and textures. This limitation is particularly critical for applications requiring high-fidelity images, such as military reconnaissance or detailed environmental monitoring. The transition from low-resolution to high-resolution translation remains an active area of research, with methods like Pix2PixHD attempting to bridge this gap [12].

**2.2 Semantic Consistency**

Another significant challenge is maintaining semantic consistency between the SAR and EO images. While cycle consistency helps, it does not guarantee that the semantic features (e.g., buildings, roads, vegetation) in SAR images will be accurately translated to their EO counterparts. This issue often results in generated images that, while visually plausible, may not be useful for practical applications that require precise object recognition and analysis.

**2.3 Data Scarcity and Diversity**

The performance of GAN models is heavily dependent on the availability and diversity of training data. SAR to EO translation suffers from a lack of large, annotated datasets that cover a wide range of scenes and conditions. This scarcity limits the ability of models to generalize well across different environments and applications. Moreover, the difference in acquisition methods for SAR and EO images can introduce alignment issues, making the training process more complex.

**2.4 Computational Complexity**

Training GANs, particularly for high-resolution image synthesis, is computationally intensive. The need for large-scale data, coupled with the complexity of the models, demands significant computational resources. This requirement poses a barrier for many research institutions and practitioners who may not have access to such resources.

**2.5 User Transparency**

While many current approaches, including the ones previously discussed, offer significant improvement in the above problem areas, there has been very little research and improvement in the transparency of image translations for the sake of end user interpretability. While there can still be large improvements in resolution details, consistency, etc., there must be a way for the end user to feel confident in the translation if they are expected to be able to utilize the translation in a realistic scenario.

## 3. METHODOLOGY

This paper proposes two novel techniques that address some of the above issues. Regarding resolution and detail preservation, we promote the use of SAR denoising and transformers to enhance the optical translation capabilities of our model as well as limiting computational complexity by working with a small training dataset. Additionally, our main contribution is in the area of user transparency whereby we introduce a novel method for increasing user confidence in translated optical images.

**3.1. Dual-Feature SAR2EO**

While many of the previously mentioned models provided significant breakthroughs and benefits each in their own ways, each also lacks the benefit of some of the advancements that the other models introduced. This research focuses on the benefit of SAR denoising/de-speckling and the utilization of transformers for improving SAR-to-EO translations in a network we're proposing as Dual-Feature SAR2EO or DF-SAR2EO.

### 3.1.1 Dataset

Our dataset was gathered from freely available Sentinel 1 and Sentinel 2 data images from the EU Copernicus Browser [13]. Each Sentinel 1 and Sentinel 2 image was captured no more than 1 day apart from each other for the sake of ensuring the images contained no significant changes. In order to decrease the computational complexity needs, the total training set consisted of 2100 256x256 SAR images. We realize than in order to develop a high-performance SAR-to-EO translation model we would need much more training data, but our goal regarding translation performance is to prove that SAR denoising and transformers provide significant improvements regardless of the amount of training data.

In order to further prepare the data for the DF-SAR2EO architecture, we augment the training data so that we ultimately have three unique data inputs for the model. For our first input, we run canny edge detection on the SAR image to extract the image edges then for our second we simply convert the RGB SAR image to grayscale. The RGB SAR image is used as our third input.

### 3.1.2 Training Pipeline

The pipeline structure of DF-SAR2EO is illustrated in Figure 4. It begins with running a SAR denoising algorithm in order to remove excess noise and speckling from the SAR images. Following that, the training data is fed into the generator consisting of two encoder-decoder sub-generators, one for texture and one for structure, similar to the dual-feature SAR-to-EO translator [6]. The data then is fed into a Bi-GFF module and then a CFA module which ultimately produces the generated image. Lastly, the generated image is again run through a canny edge detector to extract its edges and then passed through a discriminator to determine if the generated image is real or fake. Throughout the model, several loss functions are used to determine the image quality and train the model to improve the generation.

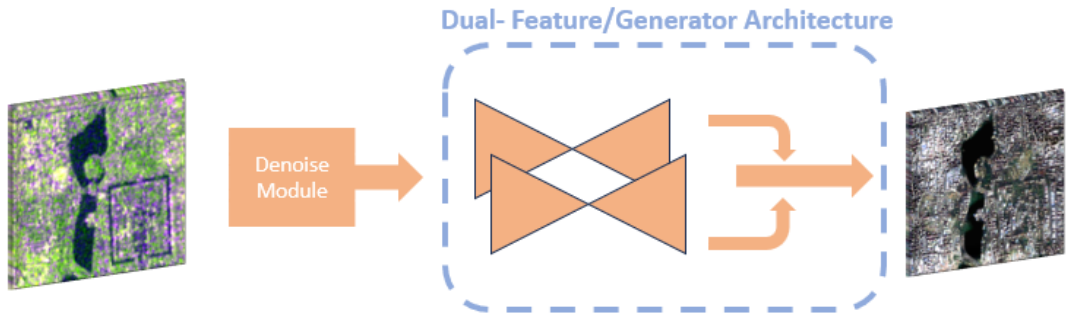

*Figure 4: DF-SAR2EO Training Pipeline*

### 3.1.3 Model Architecture

Our model begins with a dual-feature generator consisting of two sub-generators each of which is focused on extracting either a structure feature or texture feature. Our structure encoder first receives the concatenated SAR edge image and grayscale SAR image as input and then begins the process of encoding structural information. The texture encoder, on the other hand, processes the 3-channel SAR image to extract texture-related features.

*Structure Encoder.* The structure encoder utilizes partial convolution layers (PConv2d) for its operations, which helps in handling the irregular input by masking invalid regions. Specifically, it processes the input through a series of partial convolution layers with varying filter sizes and strides to progressively extract high-level structural features. The layers include PConv2d layers responsible for the initial feature extraction with progressively increasing channels followed by batch normalization applied after each convolution to stabilize and accelerate training.

*Texture Encoder.* Similarly, the texture encoder employs partial convolution layers to capture textural details from the input SAR image. The texture encoder follows the same architectural pattern as the structure encoder in addition to using partial convolutions and batch normalization.

*Structure Decoder* The structure decoder in our architecture is designed to reconstruct high-resolution structural features from the encoded features, ensuring that the global structure of the generated optical images remains consistent and accurate. This decoder utilizes a combination of deconvolutions, skip connections, and progressive upsampling to ensure detailed and accurate reconstruction. The structure decoder begins by receiving the highest-level features from the texture encoder and progressively upsamples these features through a series of deconvolution layers. Each deconvolution layer is followed by batch normalization and a LeakyReLU activation function, which helps in maintaining stable gradients and improving convergence during training. The first two layers also use dropout connections to aid in the generalization of the generated features.

At each deconvolution stage, the upsampled features are concatenated with corresponding features from the structure encoder through skip connections. These skip connections ensure that high-resolution details from the encoding phase are preserved and directly available to the decoder. This process of concatenation followed by deconvolution allows the network to combine coarse, high-level features with fine, detailed features, resulting in a more accurate reconstruction.

*Texture Decoder.* The texture decoder is focused on reconstructing the fine texture details of the generated optical images. Its architecture is exactly the same as the structure decoder, using a series of deconvolutions to progressively upsample the feature maps.

By using a combination of deconvolutions, skip connections, and upsampling, both the structure and texture decoders ensure that the generated optical images are detailed and accurate, effectively capturing both global structures and fine textures. The integration of features from both encoders at each deconvolution step ensures that the final output images are coherent and visually appealing, with a balanced representation of structural and textural details. The following tables detail the parameters used for the texture and structure encoders and decoders.

| Module Name | Filter Size | Channel | Stride | Padding | Nonlinearity |
|---|---|---|---|---|---|
| **Texture/Structure (T/S) Encoder** | | | | | |
| T/S Input | - | 3/2 | - | - | - |
| T/S Encoder PConv1 | 7 x 7 | 64 | 2 | 3 | ReLU |
| T/S Encoder PConv2 | 5 x 5 | 128 | 2 | 2 | ReLU |
| BatchNorm2d | - | 128 | - | - | - |
| T/S Encoder PConv3 | 5 x 5 | 256 | 2 | 2 | ReLU |
| BatchNorm2d | - | 256 | - | - | - |
| T/S Encoder PConv4 | 3 x 3 | 512 | 2 | 1 | ReLU |
| BatchNorm2d | - | 512 | - | - | - |
| T/S Encoder PConv5 | 3 x 3 | 512 | 2 | 1 | ReLU |
| BatchNorm2d | - | 512 | - | - | - |
| T/S Encoder PConv6 | 3 x 3 | 512 | 2 | 1 | ReLU |
| BatchNorm2d | - | 512 | - | - | - |
| T/S Encoder PConv7 | 3 x 3 | 512 | 2 | 1 | ReLU |

*Table 5: DF-SAR2EO Encoder Parameters.*

| Module Name | Filter Size | Channel | Stride | Padding | Nonlinearity |
|---|---|---|---|---|---|
| **Texture Decoder** | | | | | |
| S Encoder-PConv7 Upsampled 4x4 | - | 512 | - | - | - |
| Concat(S Encoder-PConv7 Up, T Encoder-PConv6) | - | 512 + 512 | - | - | - |
| T Decoder PConv8 | 3x3 | 512 | 1 | 1 | LeakyReLU |
| T Decoder-PConv8 Upsampled 8x8 | - | 512 | - | - | - |
| Concat(T Decoder-PConv8 Up, T Encoder-PConv5) | - | 512 + 512 | - | - | - |
| T Decoder PConv9 | 3x3 | 512 | 1 | 1 | LeakyReLU |
| T Decoder-PConv9 Upsampled 16x16 | - | 512 | - | - | - |
| Concat(T Decoder-PConv9 Up, T Encoder-PConv4) | - | 512 + 512 | - | - | - |
| T Decoder PConv10 | 3x3 | 512 | 1 | 1 | LeakyReLU |
| T Decoder-PConv10 Upsampled 32x32 | - | 512 | - | - | - |
| Concat(T Decoder-PConv10 Up, T Encoder-PConv3) | - | 512 + 256 | - | - | - |
| T Decoder PConv11 | 3x3 | 256 | 1 | 1 | LeakyReLU |
| T Decoder-PConv11 Upsampled 64x64 | - | 256 | - | - | - |
| Concat(T Decoder-PConv11 Up, T Encoder-PConv2) | - | 256+128 | - | - | - |
| T Decoder PConv12 | 3x3 | 128 | 1 | 1 | LeakyReLU |
| T Decoder-PConv12 Upsampled 128x128 | - | 128 | - | - | - |
| Concat(T Decoder-PConv12 Up, T Encoder-PConv1) | - | 128+64 | - | - | - |
| T Decoder PConv13 | 3x3 | 64 | 1 | 1 | LeakyReLU |
| T Decoder-PConv13 Upsampled 256x256 | - | 64 | - | - | - |
| Concat(T Decoder-PConv13 Up, T Encoder Input) | - | 64+3 | - | - | - |
| **Texture Feature** | 3x3 | 64 | 1 | 1 | LeakyReLU |

Table 6: DF-SAR2EO Texture Decoder Parameters.

| Module Name | Filter Size | Channel | Stride | Padding | Nonlinearity |
|---|---|---|---|---|---|
| **Structure Decoder** | | | | | |
| T Encoder-PConv7 Upsampled 4x4 | - | 512 | - | - | - |
| Concat(T Encoder-PConv7 Up, S Encoder-PConv6) | - | 512 + 512 | - | - | - |
| S Decoder PConv14 | 3x3 | 512 | 1 | 1 | LeakyReLU |
| S Decoder-PConv 14 Upsampled 8x8 | - | 512 | - | - | - |
| Concat(S Decoder-PConv14 Up, T Encoder-PConv5) | - | 512 + 512 | - | - | - |
| S Decoder PConv15 | 3x3 | 512 | 1 | 1 | LeakyReLU |
| S Decoder-PConv15 Upsampled 16x16 | - | 512 | - | - | - |
| Concat(S Decoder-PConv15 Up, T Encoder-PConv4) | - | 512 + 512 | - | - | - |
| S Decoder PConv16 | 3x3 | 512 | 1 | 1 | LeakyReLU |
| S Decoder-PConv16 Upsampled 32x32 | - | 512 | - | - | - |
| Concat(S Decoder-PConv16 Up, T Encoder-PConv3) | - | 512 + 256 | - | - | - |
| S Decoder PConv17 | 3x3 | 256 | 1 | 1 | LeakyReLU |
| S Decoder-PConv17 Upsampled 64x64 | - | 256 | - | - | - |
| Concat(S Decoder-PConv17 Up, T Encoder-PConv2) | - | 256+128 | - | - | - |
| S Decoder PConv18 | 3x3 | 128 | 1 | 1 | LeakyReLU |
| S Decoder-PConv18 Upsampled 128x128 | - | 128 | - | - | - |
| Concat(S Decoder-PConv18 Up, T Encoder-PConv1) | - | 128+64 | - | - | - |
| S Decoder PConv19 | 3x3 | 64 | 1 | 1 | LeakyReLU |
| S Decoder-PConv19 Upsampled 256x256 | - | 64 | - | - | - |
| Concat(S Decoder-PConv19 Up, S Encoder Input) | - | 64+2 | - | - | - |
| **Structure Feature** | 3x3 | 64 | 1 | 1 | LeakyReLU |

*Table 7: DF-SAR2EO Structure Decoder Parameters.*

*Bi-GFF.* The Bi-directional Global Feature Fusion (Bi-GFF) module is designed to effectively integrate the structure and texture features extracted by the respective decoders. This module aims to enhance the overall feature representation by leveraging interactions between the structure and texture features, thus ensuring that the generated optical images maintain both structural integrity and detailed texture information.

The Bi-GFF module operates by first concatenating the structure feature map and the texture feature map along the channel dimension to form a combined feature map. This concatenated feature map serves as the input to two separate convolutional layers, denoted as Ws and Wt, each with a kernel size of 3x3, padding of 1, and output channels set to 64. The convolutional layer Ws focuses on refining the concatenated features with a bias towards structure features, while the convolutional layer Wt does the same with a bias towards texture features. These convolutions transform the concatenated features into two intermediate feature maps.

Subsequently, element-wise multiplication is performed between these two feature maps and the input features. This interaction allows for a cross-enhancement of features, where the structural information enriches the texture features and vice versa. The resulting feature maps from these multiplications are then added element-wise back to their respective original features. This operation yields two enhanced feature maps which are more robust and comprehensive representations of the structure and texture, respectively.

Finally, the Bi-GFF module concatenates these enhanced feature maps along the channel dimension, resulting in a fused feature map that integrates both the refined structure and texture information. This fused feature map is then utilized in subsequent layers of the network to generate the final high-quality optical images. Figure 5 shows an illustration of the Bi-GFF module as explained by the original authors.

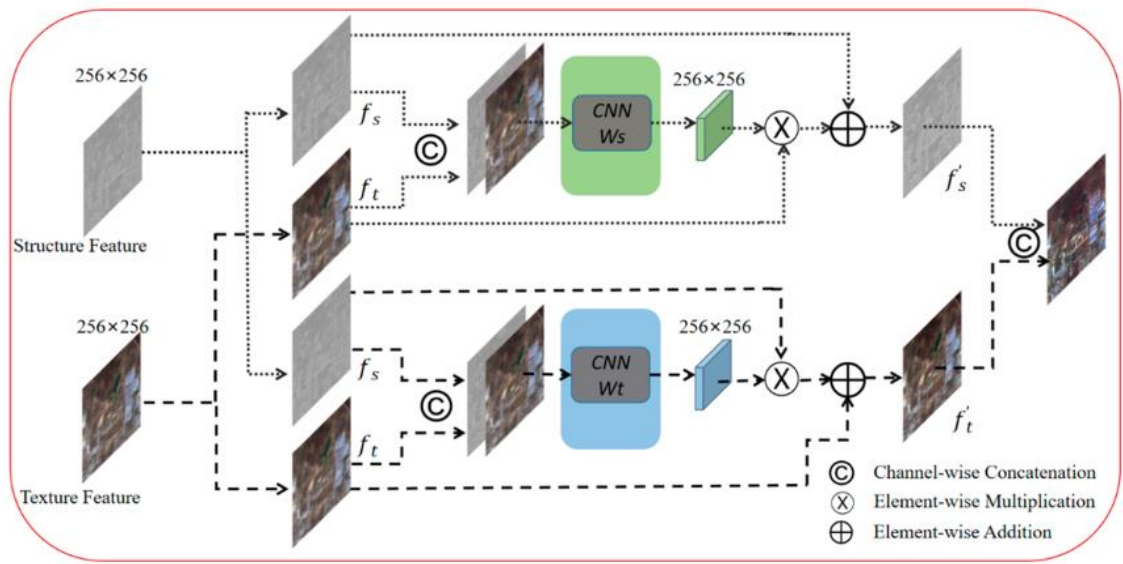

*Figure 5: Bi-GFF Module as illustrated in Nie et al.'s 2022 paper.*

*CFA*. The Cross-Feature Attention (CFA) module is designed to enhance the feature representation by utilizing attention mechanisms to selectively focus on relevant features across different scales and spatial locations. This module plays a similar role to that of a transformer, leveraging self-attention mechanisms to capture long-range dependencies and context within the feature maps. Here, we detail the architecture and functionality of the CFA module, highlighting its importance in refining the generated optical images.

The CFA module begins by applying three convolutional layers with a kernel size of 3x3 and stride of 1, each followed by a batch normalization layer, to the input feature map Fin. These initial convolutions serve to preprocess the features, making them suitable for the subsequent attention mechanism.

Next, the preprocessed feature map is divided into a series of flattened patches, facilitating the computation of attention scores. Each patch is normalized over the channel dimension to ensure consistent scale across different patches. The normalized patches are then used to compute a cosine similarity matrix, representing the similarity between every pair of patches. This similarity matrix is passed through a softmax function to obtain the attention scores, which indicate the importance of each patch relative to the others.

Using these attention scores, the feature map is reconstructed by performing a weighted sum of the patches. This reconstructed feature map retains the most relevant information from the original feature map, emphasizing important regions while suppressing less significant ones.

The CFA module further enhances the new feature map by generating multiscale features through a series of dilated convolutions with dilation rates of 1, 2, 4, and 8. These multiscale features capture context at varying spatial resolutions, allowing the network to integrate both fine-grained and coarse information. The weights for combining these multiscale features are learned through a 1x1 convolution, which outputs four separate weight tensors.

The final step involves combining the multiscale features using the learned weights. Each feature map is multiplied by its corresponding weight tensor, and the results are summed to produce the final refined feature map. This refined feature map is concatenated with the original input feature map via a skip connection, ensuring that the network retains important low-level information.

The concatenated feature map is then passed through a series of three convolutional layers with Leaky ReLU activations and batch normalization, producing the final output of the CFA module. This output is a highly refined feature map that integrates context from multiple scales and spatial locations, significantly enhancing the quality of the generated optical images. Figure 6 shows the original author illustration of the CFA module.

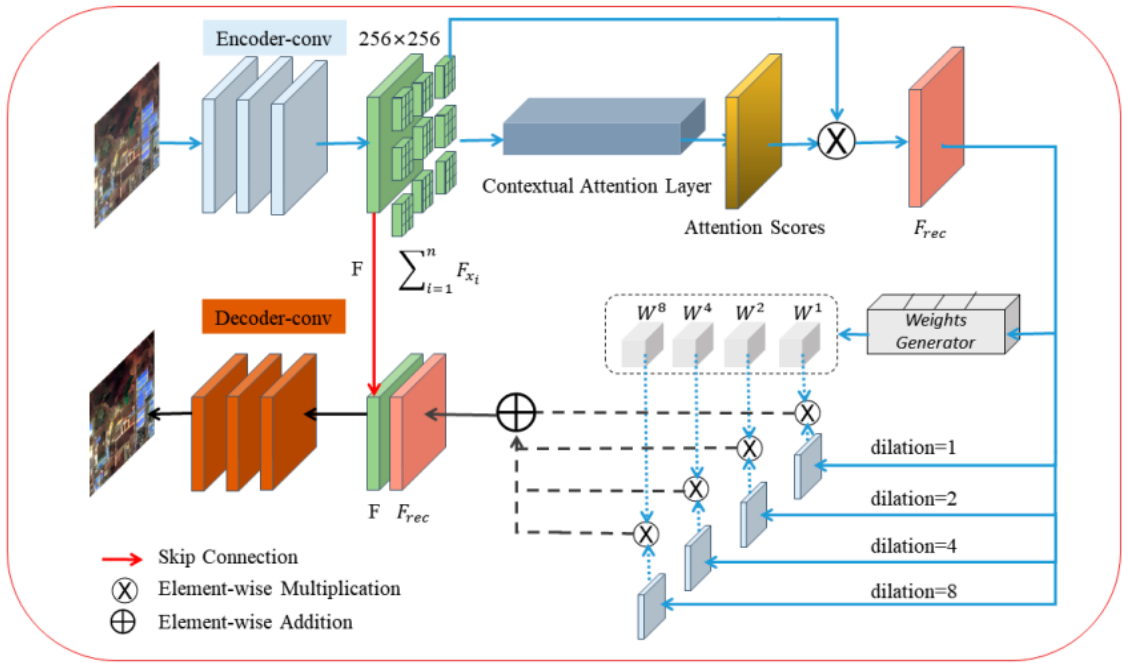

*Figure 6: CFA module as illustrated in Nie et al.'s 2022 paper.*

*Discriminator.* The discriminator employs a dual-branch architecture to effectively distinguish between pseudo-optical and real optical images by separately processing texture and structure information. The structure branch processes edge maps through a Residual Block, consisting of convolutional layers with spectral normalization, batch normalization, and LeakyReLU activations, followed by a 1x1 convolution. The texture branch mirrors this process, focusing on texture details. The structure features are then enriched by concatenating them with the grayscale version of the input images and combined with the texture features. The combined features are passed through a final convolutional layer and a sigmoid activation to produce a probability map, indicating the likelihood that each pixel in the input image belongs to a real optical image. This approach ensures a comprehensive analysis of both structural and textural characteristics, leading to accurate classification.

*Loss Functions.* The loss functions employed in our model aim to comprehensively guide the training process by addressing different aspects of image generation. The adversarial loss (BCELoss) ensures the generated images are indistinguishable from real images by the discriminator. The reconstruction losses, including MSELoss and focal frequency loss, enforce pixel-wise similarity and preserve high-frequency details, respectively. The VGG-based perceptual loss measures the similarity of feature representations between the pseudo-optical and real optical images, while the style loss, derived from Gram matrices, maintains the texture consistency. Additionally, MSE loss is used for both structure and texture features to ensure accurate reconstruction. The combination of these loss functions balances various facets of image quality and realism, driving the generator to produce high-fidelity images. Figure 7 shows an illustration of the Dual-Generator network proposed in Nie et al's 2022 paper which also shows where the loss functions we placed throughout the network. For more detail on the loss functions used, we refer the reader to Nie et al.'s original paper [6].

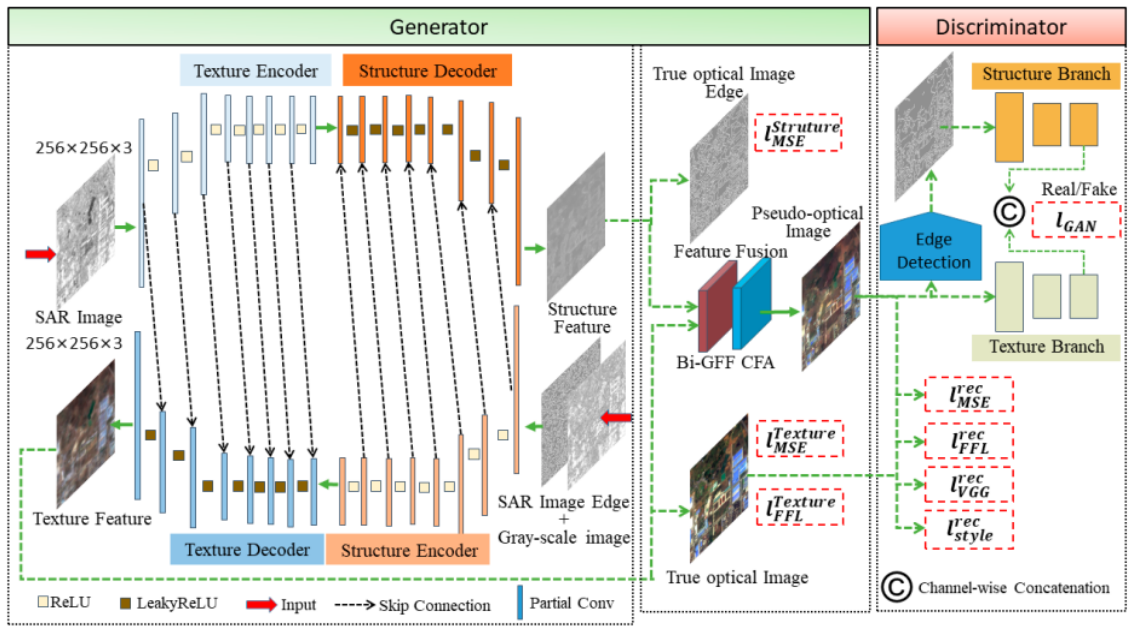

Figure 7: Dual-Generator Network Architecture from Nie et al.'s 2022 paper. Note the placement of the loss functions.

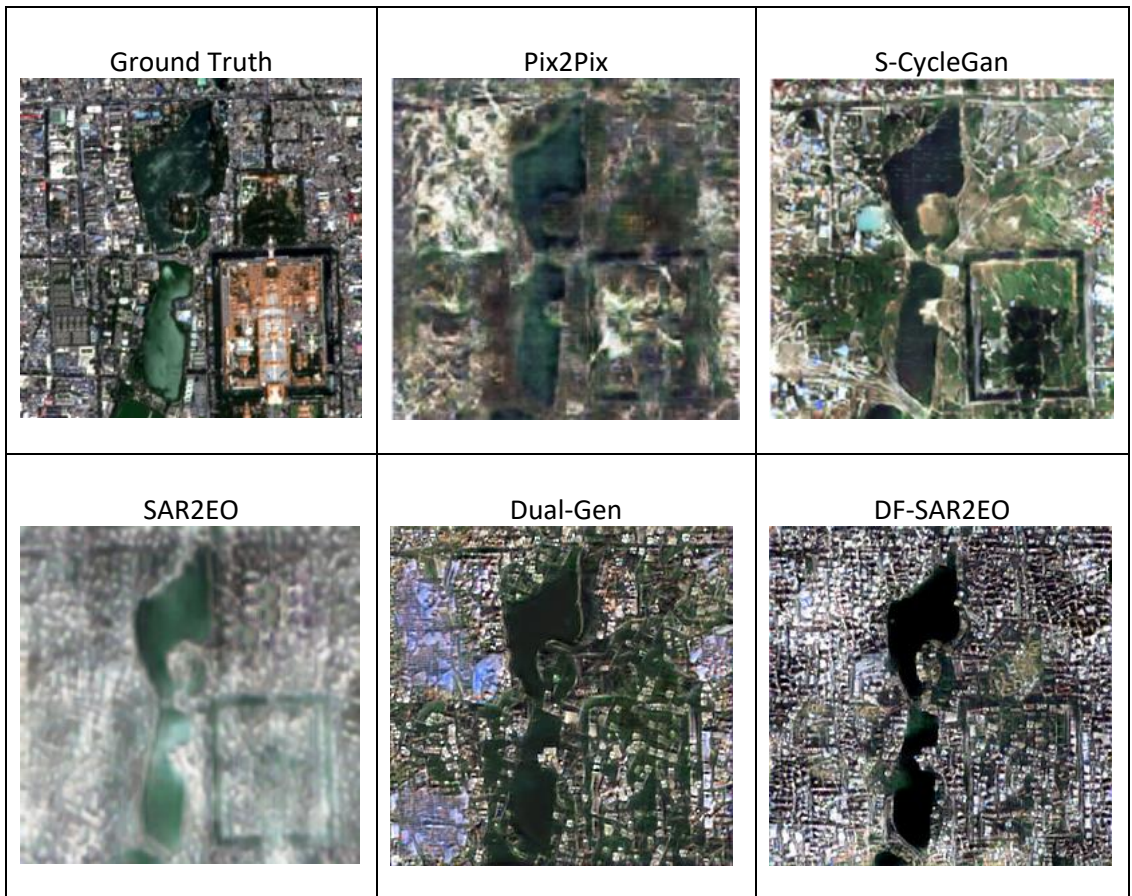

Table 8: The resulting translations of each algorithm discussed. Top left: Ground Truth. Bottom Right: Our proposed DF-SAR2EO method.

### 3.1.4 Performance Interpretability

We show in Table 8 the resulting translated optical images from each of the networks we've discussed, including our proposed DF-SAR2EO network. As stated before, the DF-SAR2EO network is one of two solutions we're proposing for the field of SAR to EO translation. In this section, we're going to focus on how the end user could engage and become more informed on the accuracy of the translated image. In a potential real-world use case for SAR to EO translation, the end user most likely will not be able to compare translations and therefore, will have no way of knowing how accurate a translation is. Therefore, we're proposing three unique metrics/visualizations to provide the end user with confidence on how a translation model is performing.

*Heatmap Overlays.* Our first interpretability method involves utilizing the probabilities produced by the sigmoid function in the discriminator. During the adversarial training process, the discriminator is trained to distinguish between real and fake images by assigning a probability score to each pixel or patch, indicating the likelihood of being real. Higher probability values correspond to higher confidence that the pixel or patch is real, while lower values indicate lower confidence. After generating a translated image, we pass this image through the discriminator, which outputs a probability map. This map is then used to construct a heatmap that visually represents the model's confidence across different regions of the image, where warmer colors (e.g., red) indicate higher confidence and cooler colors (e.g., blue) indicate lower confidence. To provide a comprehensive view, the heatmap can be overlaid on the translated image. This composite image allows users to visually assess which areas of the translation the model is more confident about and which areas might require further improvement.

*EO Confidence Score.* In order to quantitatively assess the quality of the translated optical images from SAR inputs, we employ a pre-trained Siamese network. This network is specifically designed to measure the similarity between a SAR image and its optical counterpart. By leveraging this network, we can obtain a confidence score that indicates how well the translated optical image matches the expected appearance of an actual optical image for the given SAR input.

The Siamese network comprises two identical subnetworks, each processing one of the input images (SAR and optical). The architecture of the Siamese network consists of shared convolutional layers designed to extract relevant features from the image. These are made up of a convolutional layer with 64 filters of size 5x5, followed by a ReLU activation function and a max-pooling layer, then a second convolutional layer with 128 filters of size 5x5, followed by a ReLU activation function and another max-pooling layer. The output from the convolutional layers is flattened and passed through a series of fully connected layers to generate an embedding for each input image. This output is then passed through a series of three fully connected layers consisting of 512, 256, and 128 units and the first two are followed by a ReLU activation while the last one produces an embedding vector. The embeddings from the two input images are then compared to compute a similarity score. This score reflects the network's confidence in the degree of similarity between the SAR and optical images.

We trained our Siamese network using 7,520 pairs of SAR and optical images of the same location. During training, the network learns to generate similar embeddings for image pairs that are true counterparts, and dissimilar embeddings for non-matching pairs. The loss function used for training is a contrastive loss, which encourages the network to minimize the distance between embeddings of similar images and maximize the distance between embeddings of dissimilar images. The use of a Siamese network for evaluating translated images provides a robust and quantitative method to assess translation performance.

*Spatial Consistency Graph.* Due to most SAR images being much larger than 256x256, we employ a patch technique on many of our translations where we patch the image into smaller 256x256 patches and perform translations on those, then realign them into a larger translated full resolution image. This allows us the opportunity to evaluate the quality and coherence of the translated optical images from SAR images by employing a spatial consistency assessment technique. This method ensures that the translated image patches are consistent with one another, maintaining spatial coherence across the entire image. The rationale for this approach assumes that adjacent patches in a high-quality translation should seamlessly align with minimal discrepancies along their edges.

The spatial consistency assessment technique involves comparing the edges of adjacent patches to quantify their similarity. The structural similarity index measure (SSIM) is used to calculate the similarity between the edges of two adjacent patches. SSIM is a widely used metric for measuring the similarity between two images, focusing on their luminance, contrast, and structural information. It ranges from -1 to 1, where a value of 1 indicates perfect similarity.

The edge comparison is performed for both horizontal and vertical orientations. For horizontal consistency, the right edge of the current patch is compared with the left edge of the adjacent patch to the right. For vertical consistency, the bottom edge of the current patch is compared with the top edge of the adjacent patch below. Table 9 shows each of our interpretation metrics and visualizations used on two SAR-to-EO translations using our proposed DF-SAR2EO model.

| SAR Image | Translation | Heatmap | Spatial Consistency Graph | Conf. Score |
|---|---|---|---|---|
| 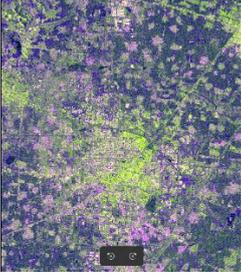 | 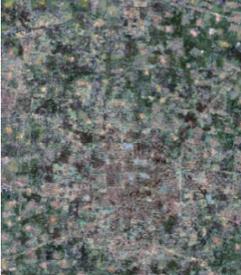 | 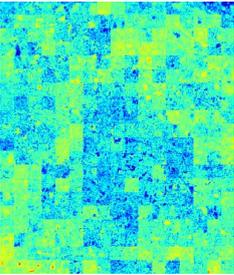 | 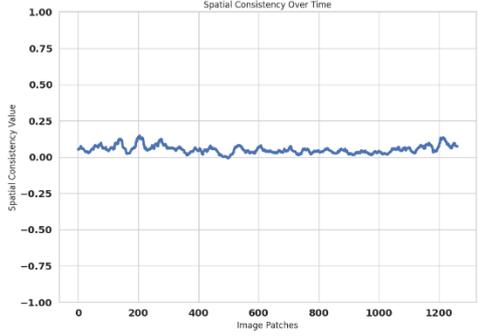 | 71.43% |
| 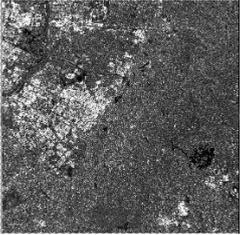 | 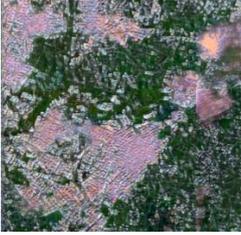 | 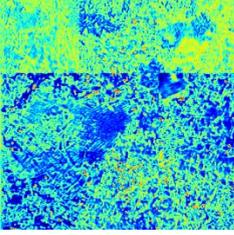 | 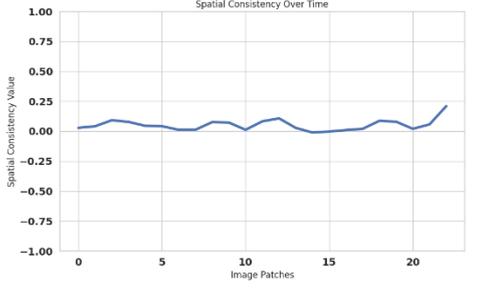 | 76.33% |
| 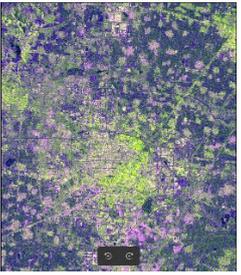 | (Ground Truth) 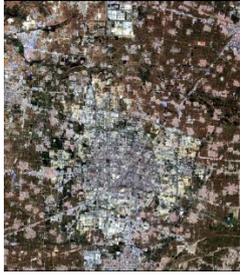 | 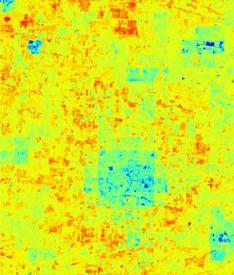 | 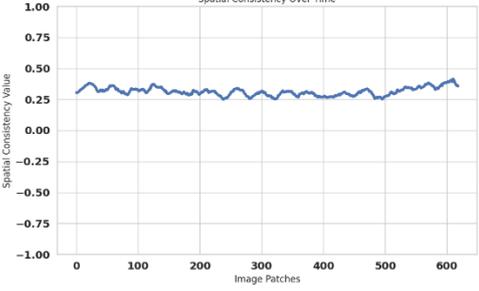 | 88.77% |

*Table 9: Our proposed visualizations and metrics for SAR-to-EO translations. Bottom row is results of transparency metrics on ground truth. Heatmap: Blue = low confidence, Red = High confidence.*

## 3. DISCUSSION

As stated previously, SAR-to-EO translation research holds significant promise for various practical applications, including environmental monitoring, urban planning, and military reconnaissance. In the realm of environmental monitoring, accurate EO image synthesis from SAR data can provide detailed insights into land use changes, deforestation, and urban expansion, which are crucial for sustainable development and environmental protection. For urban planning, synthesized EO images can aid in the assessment of infrastructure development, zoning regulations, and disaster management, providing planners with accurate and up-to-date visual information to support informed decision-making. In military reconnaissance, high-quality EO images derived from SAR data can enhance situational awareness, target identification, and mission planning, thereby improving the effectiveness and safety of operations.

Therefore, the potential impact of improved EO image synthesis on operational decision-making and planning is profound. Accurate EO images can bridge the gap between SAR and EO imagery, offering a seamless transition from one modality to another. This capability enables more robust and comprehensive analysis by leveraging the strengths of both SAR and EO data. Enhanced EO image synthesis can lead to better-informed decisions, as stakeholders can rely on high-resolution, accurate visual data for strategic planning and real-time operations. For instance, in disaster response scenarios, synthesized EO images can provide clear and actionable information on affected areas, facilitating efficient resource allocation and timely intervention.

Lastly, the effectiveness of the interpretability tools such as heatmaps, confidence scores, and spatial consistency graphs is instrumental in enhancing user trust and understanding of the model's outputs. Heatmaps offer a visual representation of the model's confidence in different regions of the synthesized image, allowing users to pinpoint areas of high and low certainty. This transparency helps users to critically evaluate the quality of the translation and make informed decisions based on the model's confidence levels. Similarly, the confidence score provides a quantitative measure of the overall accuracy of the translated image, serving as a quick reference for assessing the reliability of the output. The spatial consistency graph, which evaluates the coherence between adjacent image patches, ensures that the translated image maintains a high level of visual continuity, further reinforcing the model's credibility. These interpretability tools collectively contribute to building user trust by making the model's decision-making process more transparent and comprehensible.

## 4. FUTURE WORK

There are several avenues for future work that could further enhance the effectiveness and applicability of SAR-to-EO image translation. These directions include suggestions for improvements in GAN architectures, the integration of multimodal data sources, and the exploration of real-time translation capabilities.

## 4.1 Improvements in GAN Architectures

Future research could focus on refining GAN architectures to improve image translation quality. One potential area of exploration is the incorporation of more advanced generative models, such as StyleGAN and its variants, which have demonstrated superior performance in high-resolution image synthesis [14]. Additionally, integrating attention mechanisms within the GAN architecture could enhance the model's ability to focus on relevant features, improving the accuracy and consistency of the translated images. Exploring deeper and more complex networks, while carefully managing computational efficiency, could also yield better results. Furthermore, implementing advanced loss functions, such as perceptual loss or contextual loss, can provide more nuanced feedback during training, leading to higher fidelity translations.

## 4.2 Integration of Multimodal Data Sources

The integration of multimodal data sources presents a promising direction for enhancing translation accuracy. Combining SAR data with other remote sensing data types, such as hyperspectral imagery, could provide complementary information that enriches the translation process. Multimodal GANs, which can process and fuse information from various data sources, have the potential to produce more accurate and detailed EO images. Additionally, leveraging auxiliary information such as digital elevation models (DEMs) or weather data could further improve the contextual understanding of the scenes being translated, leading to more realistic and accurate EO image outputs.

## 4.3 Exploration of Real-Time Translation Capabilities

Developing real-time translation capabilities is another exciting area for future research. Achieving real-time performance would require optimizing the computational efficiency of the GAN models without sacrificing translation quality. Techniques such as model pruning, quantization, and efficient neural network architectures (e.g., MobileNets or EfficientNet) [15, 16] could be explored to reduce the computational burden. Real-time translation would be particularly valuable in dynamic and time-sensitive applications, such as disaster response and military reconnaissance, where timely and accurate information is crucial. Additionally, real-time capabilities could enable interactive applications, allowing users to obtain immediate feedback and make rapid decisions based on the translated EO images.

The future of SAR-to-EO image translation research is rich with potential. By focusing on improving GAN architectures, integrating multimodal data sources, and developing real-time translation capabilities, future work can significantly advance the state-of-the-art in image translation, leading to more accurate, efficient, and practical solutions for a wide range of applications.

## ACKNOWLEDGMENTS

The authors would like to thank the PeopleTec Technical Fellows program for its encouragement and project assistance.

**Authors**

| | |
|---|---|
| **Grant Rosario** has research experience in embedded applications and autonomous driving applications. He received his Masters from Florida Atlantic University in Computer Science and his Bachelors from Florida Gulf Coast University in Psychology. | 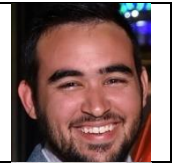 |
| **David Noever** has research experience with NASA and the Department of Defense in machine learning and data mining. He received his BS from Princeton University and his Ph.D. from Oxford University, as a Rhodes Scholar, in theoretical physics. | 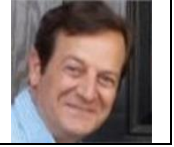 |